%% file: main.tex
\documentclass{article} 
\usepackage{collas2026_conference,times}
\usepackage{easyReview}
\input{math_commands.tex}

\usepackage{graphicx}
\usepackage{booktabs}
\usepackage{xcolor}

\usepackage{hyperref}
\hypersetup{
    colorlinks=true,
    linkcolor=red,
    filecolor=magenta,
    urlcolor=blue,
    citecolor=purple,
    pdftitle={Muon Can Outperform Dedicated Continual Learning Methods},
    pdfpagemode=FullScreen,
    }
\title{Muon Can Outperform Dedicated Continual Learning Methods}
\author{Sebastian George Sincari\thanks{Equal contribution.} \\
Faculty of Mathematics and Computer Science\\
University of Bucharest\\
Bucharest, Romania \\
\texttt{sebastian-george.sincari@s.unibuc.ro} \\
\And
Bogdan Alexandru Gheorghe\footnotemark[1] \\
Faculty of Mathematics and Computer Science\\
University of Bucharest\\
Bucharest, Romania \\
\texttt{bogdan-alexandru.gheorghe@s.unibuc.ro} \\
\And
Antonio Barbalau \\
Bitdefender\\
Bucharest, Romania \\
\texttt{abarbalau@bitdefender.com} \\
}

\collaswipcopy
\begin{document}
\maketitle

\begin{abstract}
Continual learning with Low-Rank Adapters (LoRA) typically mitigates
forgetting by penalizing the overlap between a new update and the accumulated
past weights, which discourages certain update directions without controlling
how an update distributes its energy over the ones that remain.
We ask whether that restriction has to be task-aware, or whether a generic one
supplied by the optimizer is enough. We train a plain incremental LoRA (IncLoRA)
with Muon, which orthogonalizes each update, and compare it against O-LoRA and
ELLA over five seeds and three task orders on the Standard CL Benchmark and
three seeds on TRACE. IncLoRA${+}$Muon reaches the accuracy band of the
dedicated methods on Standard CL and improves on every AdamW configuration on
TRACE.
One update-constraining mechanism is enough, whether it comes from the loss
or from the optimizer; on Standard CL a second one does not help, and for
the most restrictive method it costs $8.4$ points of accuracy and the plasticity
to fit each task.
What separates the two optimizers is not the size of the update, which under
Muon is $0.91$ to $2.06\times$ that under AdamW, but how it is distributed.
AdamW confines it to between $1.4$ and $1.8$ effective singular directions, Muon
spreads it over $7.0$, and the two do not overlap in any tracked run.
Part of the advantage usually attributed to dedicated CL methods may
therefore be explained by the geometry of the optimizer's updates.
\end{abstract}

\section{Introduction}

Continually fine-tuning a pretrained model on a sequence of tasks remains
difficult because of \emph{catastrophic forgetting}: adapting to a new task
tends to overwrite the representations learned for previous ones. In the
large-model regime, full fine-tuning is rarely practical, so most recent work
builds on parameter-efficient fine-tuning, and on LoRA in particular
\citep{hu2022lora}. A family of continual-learning methods then augments LoRA
with mechanisms that explicitly protect previously acquired knowledge.

Two representative instances are O-LoRA \citep{wang2023olora}, which penalizes
overlap between a new update and the subspaces of earlier tasks, and ELLA
\citep{ella}, which penalizes overlap with the accumulated past weights.
Both discourage certain directions; neither controls how an update
distributes its energy over the ones that remain. Such a penalty can act on the update's magnitude as well as on its
distribution across singular directions. We measure both: the penalties act on
the first, the optimizer on the second.
This motivates our central question. If a restriction on the update is
the operative ingredient, does it have to be task-aware (computed from the
subspaces of previous tasks), or can a generic restriction be supplied at the
level of the optimizer, at no cost in machinery?

To test this, we keep the model plain, a standard IncLoRA with no CL-specific
penalty, and instead change how the update is computed, using Muon
\citep{muon}, an optimizer that orthogonalizes and normalizes each update by
construction. Under a controlled comparison protocol, we find that
IncLoRA${+}$Muon matches or exceeds ELLA and O-LoRA.
Because the differences between the best configurations turn out to be small
relative to seed-level noise, we do not rest the paper on a ranking. Our
strongest evidence is instead mechanistic: what separates the two optimizers
completely is not the size of the update but how it is distributed across
singular directions.

\paragraph{Contributions.} This work makes three contributions:
\begin{itemize}

    \item We test whether the mitigation of catastrophic forgetting can be obtained from the optimizer alone,
    by training IncLoRA with Muon (Section~\ref{sec:methods}).
    \item We show that one update-constraining mechanism is enough and
    that a second one does not add accuracy, we separate the two failure modes
    this produces by decomposing overall accuracy exactly into task fit and
    forgetting, and we rule out step size as the explanation by measuring the
    realized update magnitudes (Section~\ref{sec:experiments}).
    \item We measure the update geometry directly and find that Muon
    distributes each update over $7.0$ effective singular directions against
    AdamW's $1.4$ to $1.8$, with no overlap between the two, confirming the
    secondary hypothesis of Section~\ref{sec:methods}
    (Section~\ref{sec:experiments}).
\end{itemize}

\section{Background}
\label{sec:background}

\paragraph{Continual learning with LoRA.} We consider the standard setting in
which a pretrained model is adapted to a sequence of tasks $T_1, \dots, T_K$
seen one at a time. For each task a LoRA adapter is trained and then merged into
the frozen backbone before moving to the next task.
Let $R_{i,j}$ be the accuracy on task $T_j$ measured after training has
finished on task $T_i$, and let $b_j$ be the zero-shot accuracy of the backbone
on $T_j$. We report
{\footnotesize
\begin{equation}
\label{eq:metrics}
\mathrm{OA} = \tfrac{1}{K}\sum_{j} R_{K,j},\quad
\mathrm{fit} = \tfrac{1}{K}\sum_{j} R_{j,j},\quad
\mathrm{BWT} = \tfrac{1}{K-1}\sum_{j<K}\big(R_{K,j}-R_{j,j}\big),\quad
\mathrm{FWT} = \tfrac{1}{K-1}\sum_{j>1}\big(R_{j-1,j}-b_j\big)
\end{equation}
}that is, overall accuracy at the end of the sequence (OA, higher is better),
task \emph{fit}, the mean accuracy on each task measured immediately after
training on it and before any subsequent task can degrade it, backward transfer
(BWT, values closer to zero indicate less forgetting) and forward transfer
(FWT). Fit and BWT are an exact reparametrization of OA,
$\mathrm{OA} = \mathrm{fit} + \tfrac{K-1}{K}\,\mathrm{BWT}$, rather than a third
independent measurement. We report them because the
decomposition separates two failure modes that produce the same OA: learning
each task poorly, and learning it well and then forgetting it.
Throughout, our reference benchmarks are the Standard CL Benchmark
\citep{zhang2015character} ($K=4$) and TRACE \citep{wang2023trace}
($K=8$), a more challenging benchmark that provides a longer and more
diverse sequence of tasks.

\paragraph{Muon.} Muon \citep{muon} is an optimizer for the weight matrices of a
network that replaces each update of the raw gradient by an (approximately) orthogonal
one, obtained through a few Newton-Schulz iterations. Intuitively, the resulting
update has its singular values normalized toward one, which means Muon directly
controls the geometry and the scale of each step \emph{by construction}, rather
than through an auxiliary loss term.
Fixing the scale is not the same as reducing it: as
Table~\ref{tab:magnitude} shows, Muon does not reduce the realized update norm
relative to AdamW on our Standard CL runs. One consequence is worth stating in
advance, since it is visible already in the construction: Muon discards the
magnitude of the momentum matrix and retains only its orthogonalized direction,
so a penalty added to the loss can still influence \emph{where} an update points
but no longer \emph{how far} it goes.

\section{Methods}
\label{sec:methods}

\paragraph{Hypothesis.} Standard CL methods mitigate forgetting by explicitly
constraining updates to minimize interference with past tasks. We
hypothesize that what matters is that the update is constrained at all, not that
the constraint is task-aware, and that Muon's orthogonalized steps can therefore
substitute for an explicit penalty. Our secondary hypothesis concerns the mechanism. AdamW
\citep{loshchilov2019adamw} rescales each coordinate on its own, while Muon acts
on the whole gradient matrix at once \citep{lau2025polargrad} and by
construction equalizes the singular values of the matrices it updates
\citep{muon}. That guarantee applies to the LoRA factors, not to their product,
whose rank is not bounded by theirs; we hypothesize that it carries over, so
that AdamW concentrates each merged update in a few dominant directions while
Muon spreads it across many. We test the first hypothesis by comparing accuracy
across configurations grouped by how many mechanisms restrict the update, and
the second by measuring the concentration of each merged update per training
step (Section~\ref{sec:experiments}).

\paragraph{Measuring update concentration.} For an update $\Delta W$ with
singular values $\sigma_1 \ge \dots \ge \sigma_\rho$ we use the ratio
$\sigma_1/\lVert \Delta W \rVert_F \in [1/\sqrt{\rho},\, 1]$, which equals $1$
when all the energy of the update lies in a single singular direction and
$1/\sqrt{\rho}$ when the spectrum is flat across $\rho$ of them. Its reciprocal
square is the \emph{stable rank}: an effective count of the directions the
update actually uses. We evaluate it on the merged update
$\Delta W_t = B_t A_t - B_{t-1} A_{t-1}$ between consecutive optimizer steps of
the adapter currently being trained, per module, and take the median across
modules (Appendix~\ref{app:concentration}). The ratio is dimensionless, so it is
unaffected by the LoRA scaling $\alpha/r$ and by any learning-rate choice.

Because $\Delta W_t$ is a difference of two rank-$r$ products its rank is at
most $2r$, so the ratio is bounded below by $1/\sqrt{2r} = 0.25$ for $r = 8$.
Empirically the spectrum is effectively of rank $r$
(Appendix~\ref{app:concentration}), so we take $1/\sqrt{r} = 0.354$, that is
$8$ directions, as the flat-spectrum reference; it is also the value Muon's
construction imposes on the update to each factor.

\paragraph{Methods compared.} We compare three continual-learning methods, IncLoRA, O-LoRA, and ELLA, each trained with both AdamW and Muon, yielding six method-optimizer combinations in total. We evaluate every combination in two distinct settings: (i) the Standard CL Benchmark (across task Orders 1, 2, and 3 from the ELLA protocol), where the tasks are relatively well-aligned text classification benchmarks; and (ii) the TRACE Benchmark (corresponding to Order 7 in the ELLA protocol). Unlike the Standard Benchmark, TRACE features a highly diverse sequence of tasks spanning code, multilingual data, and reasoning. \footnote{The eight
TRACE tasks are C-STANCE \citep{zhao2023cstance}, FOMC \citep{shah2023trillion},
MeetingBank \citep{hu2023meetingbank}, Py150 \citep{lu2021codexglue}, ScienceQA
\citep{lu2022scienceqa}, NumGLUE-cm and NumGLUE-ds \citep{mishra2022numglue},
and 20Minuten \citep{rios202120minuten}.} This higher distribution shift and inherent cross-task orthogonality allow us to evaluate how both optimizers and CL constraints behave under severe domain changes.

\paragraph{Protocol.} Appendix~\ref{app:protocol} summarizes the two settings.
Both use LoRA rank $8$ with $\alpha = 16$ on the query and value projections
\citep{hu2022lora}; since \citet{ella} specify the rank but not $\alpha$, we
swept $16$, $32$ and $64$ and selected $16$, which most closely recovered the
reported baseline. Within each setting all methods share the same data, task
order, rank, scaling and stopping criterion, so no method is advantaged by a
longer training budget. Following \citet{ella}, Standard CL uses T5-Large
\citep{raffel2020t5} with the hyperparameters of that paper; TRACE uses
Flan-T5-Large \citep{chung2024flant5}.

The two settings differ in one respect that matters for how the results should
be read. On TRACE we add two controls the published protocol does not have: we
equalize the nominal per-entry update magnitude across optimizers by the rule
below, and we replace the fixed epoch budget with early stopping on
a held-out validation loss.
On Standard CL we keep the published protocol, both its fixed epoch budget
and its learning rate, so that our numbers remain comparable with the ones
reported there; both optimizers run at the same learning rate and the matching
rule is \emph{not} applied. That comparison therefore varies the nominal scale
of the update as well as its direction, so we measure the \emph{realized}
magnitudes (Section~\ref{sec:experiments}) and find that Muon's are not the
smaller ones, which is what a scale-based explanation would require. The
concentration measurement is dimensionless and is unaffected either way.

The equalization on TRACE is motivated by two observations. Empirically, at the
AdamW learning rate Muon stopped at a visibly higher training loss than AdamW
under that criterion, which we read as too small an effective step.
Theoretically, \citet{liu2025muonscalable} derive a per-parameter update-scale
rule matching Muon's effective update magnitude to AdamW's, which for an adapter
update of shape $m \times n$ gives
\[
    \eta_{\text{Muon}} \;=\; \eta_{\text{AdamW}} \,\cdot\, \sqrt{\max(m,n)},
\]
with AdamW as the anchor (derivation in
Appendix~\ref{app:scaling-equivalence}).

The derivation carries an $\mathcal{O}(1)$ constant $c$ that we set to one
without verifying it (Appendix~\ref{app:scaling-equivalence}).

\section{Results}
\label{sec:experiments}

\paragraph{Metrics.} We report forward transfer (FWT), backward transfer (BWT),
task fit, and overall accuracy (OA), all as defined in
Equation~\ref{eq:metrics}; recall that for BWT values closer to zero indicate
less forgetting. Table~\ref{tab:trace} reports the TRACE comparison and
Table~\ref{tab:sc} the Standard CL comparison; in both, every method is trained
with both AdamW and Muon, and every entry is a mean over all runs of that
configuration with the standard deviation across runs.
Table~\ref{tab:magnitude} reports the realized update magnitudes and final
training losses, Table~\ref{tab:concentration} the update-concentration
measurement, and Table~\ref{tab:paired} the optimizer effects.
Those last are reported as \emph{paired} differences, because the task order
alone moves a configuration's mean OA by up to $6.1$ points, which is larger
than most of the differences between methods: within each
$(\text{seed}, \text{order})$ cell we take Muon $-$ AdamW for the same method,
so the order effect cancels exactly rather than entering the error term.
Alongside each mean difference we give how many of the $15$ (or $3$) cells share
its sign, which with this many runs is more informative than a standard error.

\paragraph{Our pipeline reproduces ELLA, and our unconstrained baseline is
higher than the one reported.} On Standard CL Benchmark with AdamW
(Table~\ref{tab:sc}), our ELLA run reaches an OA of $79.64$, matching the
$79.9$ reported by \citet{ella} to within seed-level variation and
confirming that our pipeline behaves as intended. On the same setting, our
IncLoRA baseline, which stacks adapters with no continual-learning constraint,
reaches $75.83$, above the $63.6$ reported for IncLoRA by \citet{ella}.
We cannot say what accounts for the difference; the configuration we use is the
one described in Section~\ref{sec:methods}, and IncLoRA with AdamW is also the
highest-variance configuration we ran ($\text{SD} = 4.33$ across the $15$ runs),
so this baseline is the one most sensitive to the choice of seed.

\begin{figure}[t]
    \centering
    \includegraphics[width=1\textwidth]{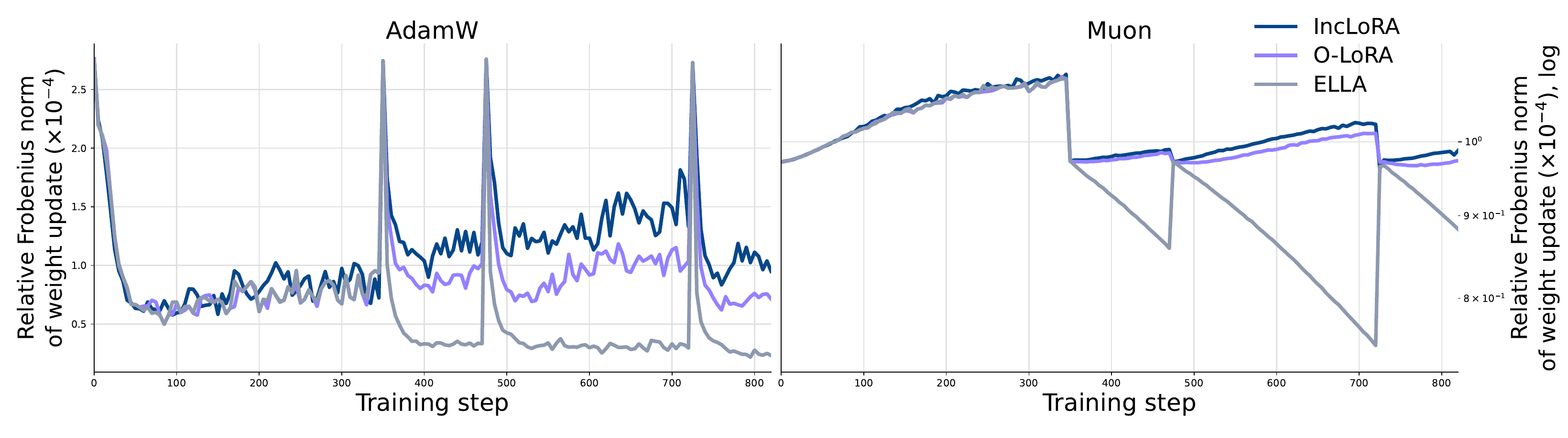}
    \caption{
    \footnotesize
    Relative Frobenius norm of the weight update
    (Appendix~\ref{formula:frobenius-norm}) during training, three methods under
    both optimizers, Standard CL Order~1. The vertical axes differ: the
    AdamW panel is linear over $0.25$ to $2.75$, the Muon panel logarithmic over
    $0.8$ to $1.0$, so the structure visible on the right is small in absolute
    terms}. Up to the first task
    boundary the penalties have no accumulated past to act on and the three curves
    coincide. Afterwards AdamW separates them by magnitude, while under Muon they
    stay together at a level the AdamW curves straddle: the orthogonalized step
    fixes the norm, so a penalty influences where the update points rather than how
    far it travels. Spikes mark task boundaries. Medians in
    Table~\ref{tab:magnitude}.
    \label{fig:update_magnitude}
\end{figure}

\paragraph{The differences are not a step-size effect.} On Standard CL both
optimizers run at the same learning rate (Section~\ref{sec:methods}), so we
measured whether their steps differ in practice. Muon's updates are not the
smaller ones: the median Muon/AdamW ratio is $0.91$ for IncLoRA, $1.18$ for
O-LoRA and $2.06$ for ELLA (Table~\ref{tab:magnitude}). Training loss is higher
under Muon for all three methods ($1.19\times$, $1.58\times$, $2.62\times$), but
for IncLoRA and O-LoRA this does not appear as worse task learning (fit
$83.1 \to 83.3$ and $82.3 \to 82.4$), so it is reached without a loss of
accuracy on the task just trained. Undertraining is therefore not an available
explanation for the retention results below, except for ELLA, whose fit does
drop.

\paragraph{One update-constraining mechanism is enough; a second one does not
add.} On Standard CL, with no mechanism restricting the update
(IncLoRA${+}$AdamW) OA is $75.83$. With exactly one, whether it comes from the
loss or from the optimizer, the three configurations land in a band of $1.6$
points: IncLoRA${+}$Muon $81.24$, O-LoRA${+}$AdamW $81.00$, ELLA${+}$AdamW
$79.64$. With two, the outcome depends on how much the explicit constraint
already restricts: O-LoRA${+}$Muon falls inside that band ($80.28$) and gains
nothing over either mechanism alone, while ELLA${+}$Muon falls $8.4$ points
below its bottom ($71.24$). The source of the mechanism matters less than
whether there is one, which is the sense in which optimizer geometry
substitutes for an explicit CL constraint. On TRACE the grouping carries no
information: there the optimizer alone separates the six, all three Muon
configurations sitting above all three AdamW ones, and the nominally best
configuration has two mechanisms (O-LoRA${+}$Muon $31.80$ against
IncLoRA${+}$Muon's $31.36$, $\text{SD} = 1.81$). Neither benchmark supports a
ranking among the leaders: on Standard CL IncLoRA${+}$Muon's margin over
O-LoRA${+}$AdamW is $0.24$ points and it wins in $7$ of $15$ paired cells.

\paragraph{Fit and forgetting separate the two failure modes.} The decomposition
in Table~\ref{tab:sc} shows that five of the six configurations fit each task
within $[80.7, 83.3]$, so no method learns a task better than the others. Two
configurations nonetheless fall short of the leaders, and on different
coordinates. IncLoRA${+}$AdamW fits as well as any ($83.1$) but has the BWT
furthest from zero ($-9.70$ against $-2.85$ for the next one): its entire
deficit is forgetting. Muon removes that deficit without touching fit ($83.3$,
BWT $-2.73$) and cuts the spread across seeds and orders by a third
($\text{SD}$ $4.33 \to 2.80$). ELLA${+}$Muon is the mirror image: it is the one
configuration outside the fit band, at $72.7$, while its retention is in line
with the rest ($-1.93$). It does not forget, because it learns less. This is not
a step-size effect either, and runs opposite to what one might expect:
ELLA${+}$Muon takes the largest realized updates of any ELLA configuration
and still terminates at the highest training loss
(Table~\ref{tab:magnitude}). The two mechanisms appear to interact
rather than to add.%

\paragraph{The more a method already restricts the update, the less Muon
adds.} On Standard CL, switching from AdamW to Muon improves IncLoRA by
$+5.41$ OA, leaves O-LoRA essentially unchanged ($-0.72$) and
degrades ELLA by $-8.40$ (Table~\ref{tab:paired}).
That ordering follows the ordering of the AdamW update magnitudes of
Table~\ref{tab:magnitude}: the method whose updates the penalty leaves untouched
gains the most, and the one whose updates it reduces most loses the most. On
TRACE Muon instead improves all three methods, by $+5.44$, $+7.62$ and $+3.42$
OA, unanimously across the three seeds. With tasks as dissimilar as TRACE's
there is little interference for a constraint to prevent, so stacking one on top
of Muon costs less than it does on the well-aligned Standard CL tasks. What
holds on both benchmarks is only the position of ELLA: it gains least from Muon
on TRACE and is the only method Muon harms on Standard CL.

\begin{figure}[t]
    \centering
    \includegraphics[width=1\textwidth]{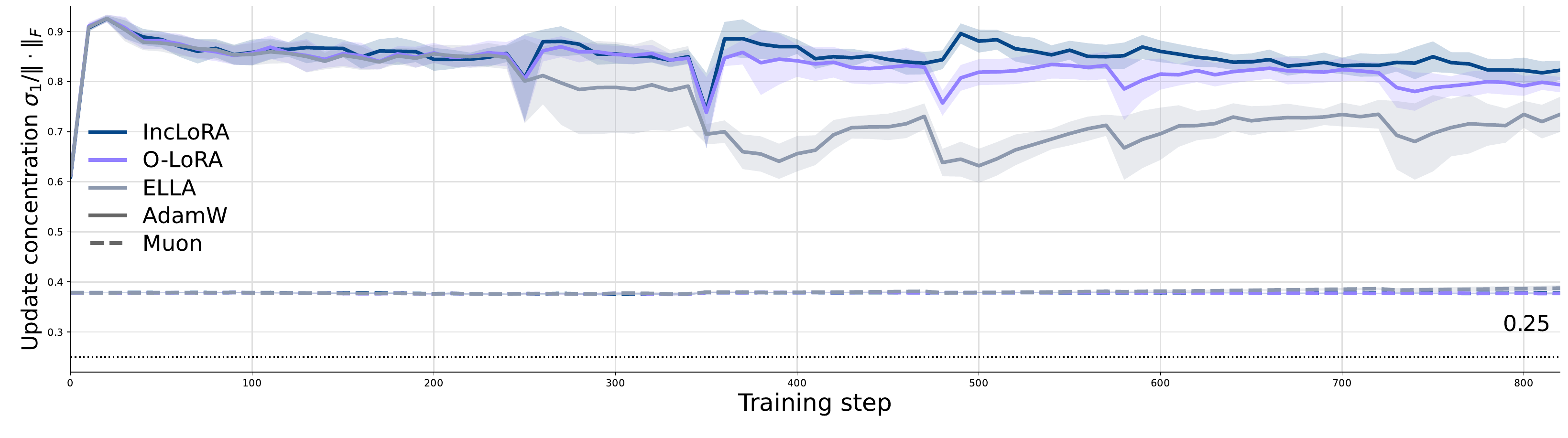}
    \caption{
    \footnotesize
    Update concentration $\sigma_1/\lVert \Delta W \rVert_F$
     per logged optimizer step, all six
    configurations, Standard CL Benchmark. Solid lines are AdamW, dashed
    Muon; bands are $\pm 1$ standard deviation across the $15$ runs of each
    configuration, which span three task orders, so the dips are not aligned.
    The dotted line is the rank-$2r$ lower bound $1/\sqrt{2r} = 0.25$ of
    Section~\ref{sec:methods}, not the flat-spectrum reference $0.354$.}
    The AdamW and Muon bands do not overlap.
    \label{fig:concentration}
\end{figure}

\paragraph{Muon spreads each update across many singular directions; AdamW
concentrates it in one or two.} Under AdamW the median concentration ratio is
$0.854$ for IncLoRA, $0.838$ for O-LoRA and $0.737$ for ELLA, that is $1.37$,
$1.42$ and $1.84$ effective directions (Table~\ref{tab:concentration}). Under
Muon all three sit at $0.378$ to $0.379$, or $6.95$ to $6.99$ directions, close
to the flat-spectrum reference of $8$ effective directions from
Section~\ref{sec:methods}. The two groups do not overlap at the decile level
anywhere: across all $90$ runs and all $83$ logged steps per run, the lowest
AdamW $10$th percentile is $0.643$ and the highest Muon $90$th percentile is
$0.386$. Neither is the separation an artefact of task transitions
(Appendix~\ref{app:concentration}). This confirms the secondary hypothesis of
Section~\ref{sec:methods}: the flat spectrum survives the product of the two
LoRA factors. Figure~\ref{fig:concentration} shows the per-step trajectories.

\paragraph{Muon moves forward transfer away from zero, in whichever
direction the sequence already points.}
The forward-transfer results reverse sign between the two benchmarks. On
Standard CL, where the tasks are well-aligned text classification problems,
Muon raises FWT by $+18.19$ points for IncLoRA and $+18.30$ for O-LoRA, in
$15/15$ paired cells for both, and by $+9.94$ for ELLA (in $10/15$). On TRACE,
where the tasks span code, multilingual data and reasoning, it pushes FWT
further negative for all three ($-2.93$, $-3.46$, $-1.87$), in $3/3$ seeds each.
Muon therefore moves FWT further from zero, keeping the sign the configuration
already had under AdamW; since the zero-shot baseline cancels term by term in a
paired difference, this holds regardless of the absolute level of FWT. ELLA is
the smaller mover on both benchmarks, at a little over half the other two: the
method that restricts the update most is the hardest to move off zero in either
direction.%
Read together with the concentration result, one reading is that an update
distributed across many directions transfers to the extent the tasks permit,
positively when they are similar and negatively when they are not. We offer it
as an interpretation only: the alignment of each sequence is taken from the
benchmark descriptions and not measured by us, and the movement away from zero
does not depend on it.%

Retention on TRACE points the other way. There $\lvert\text{BWT}\rvert$ is
under one point for every configuration and does not shrink under Muon
(Table~\ref{tab:trace}), so the OA gains on that benchmark come from fit rather
than from retention, unlike on Standard CL.%

\begin{table}[t]
\centering
\footnotesize
\caption{\footnotesize TRACE Benchmark (Order~7), Flan-T5-Large. Forward transfer (FWT),
backward transfer (BWT, closer to zero is better), task fit and overall accuracy
(OA). Mean $\pm$ standard deviation over three seeds. Best OA per optimizer in
bold.}
\label{tab:trace}
\begin{tabular}{lcccc cccc}
\toprule
& \multicolumn{4}{c}{AdamW} & \multicolumn{4}{c}{Muon} \\
\cmidrule(lr){2-5} \cmidrule(lr){6-9}
& FWT & BWT & fit & OA & FWT & BWT & fit & OA \\
\midrule
IncLoRA & $-2.55$ & $-0.10$ & $26.0$ & $\mathbf{25.92}$\tiny$\pm0.30$
        & $-5.48$ & $-0.93$ & $32.2$ & $31.36$\tiny$\pm0.52$ \\
O-LoRA  & $-1.98$ & $-0.23$ & $24.4$ & $24.18$\tiny$\pm0.72$
        & $-5.44$ & $-0.42$ & $32.2$ & $\mathbf{31.80}$\tiny$\pm1.81$ \\
ELLA    & $-2.47$ & $-0.20$ & $25.1$ & $24.96$\tiny$\pm0.39$
        & $-4.34$ & $-0.20$ & $28.6$ & $28.38$\tiny$\pm0.40$ \\
\bottomrule
\end{tabular}
\end{table}

\begin{table}[t]
\centering
\footnotesize
\caption{\footnotesize Standard CL Benchmark, T5-Large}. Per-order accuracy, overall
accuracy (OA, average over Orders~1 to 3), backward transfer and task fit. Mean
$\pm$ standard deviation over five seeds; OA aggregates all $15$ runs. Best OA
per optimizer in bold. The left column counts the mechanisms restricting the
update, either a penalty in the loss or orthogonalization in the optimizer. It
is a description of the configuration, not a prediction of its accuracy.
\label{tab:sc}
\begin{tabular}{llccc ccc}
\toprule
Mech. & & Order 1 & Order 2 & Order 3 & OA & BWT & fit \\
\midrule
none & IncLoRA${+}$AdamW & $78.45$\tiny$\pm4.27$ & $72.39$\tiny$\pm2.53$ & $76.63$\tiny$\pm4.10$ & $75.83$\tiny$\pm4.33$ & $-9.70$ & $83.1$ \\
\midrule
one  & IncLoRA${+}$Muon  & $82.83$\tiny$\pm1.86$ & $82.79$\tiny$\pm1.52$ & $78.09$\tiny$\pm1.75$ & $\mathbf{81.24}$\tiny$\pm2.80$ & $-2.73$ & $83.3$ \\
one  & O-LoRA${+}$AdamW  & $81.30$\tiny$\pm1.99$ & $81.49$\tiny$\pm1.87$ & $80.22$\tiny$\pm1.14$ & $\mathbf{81.00}$\tiny$\pm1.68$ & $-1.69$ & $82.3$ \\
one  & ELLA${+}$AdamW    & $80.99$\tiny$\pm0.86$ & $80.58$\tiny$\pm1.02$ & $77.35$\tiny$\pm2.80$ & $79.64$\tiny$\pm2.36$ & $-1.44$ & $80.7$ \\
\midrule
two  & O-LoRA${+}$Muon   & $81.40$\tiny$\pm1.74$ & $82.55$\tiny$\pm0.83$ & $76.88$\tiny$\pm1.84$ & $80.28$\tiny$\pm2.90$ & $-2.85$ & $82.4$ \\
two  & ELLA${+}$Muon     & $71.49$\tiny$\pm2.68$ & $71.56$\tiny$\pm3.05$ & $70.66$\tiny$\pm2.26$ & $71.24$\tiny$\pm2.52$ & $-1.93$ & $72.7$ \\
\bottomrule
\end{tabular}
\end{table}

\begin{table}[t]
\centering
\footnotesize
\caption{\footnotesize Realized update magnitude and convergence on Standard CL, where
both optimizers use the same learning rate. Magnitude is the relative Frobenius
norm of Appendix~\ref{formula:frobenius-norm}, median over logged steps then
over runs; loss is that of the last task. All $90$ runs had per-step
logs.}
\label{tab:magnitude}
\begin{tabular}{lccc ccc}
\toprule
& \multicolumn{3}{c}{magnitude ($\times 10^{-4}$)} & \multicolumn{3}{c}{final training loss} \\
\cmidrule(lr){2-4} \cmidrule(lr){5-7}
& AdamW & Muon & ratio & AdamW & Muon & ratio \\
\midrule
IncLoRA & $1.104$ & $0.999$ & $0.91\times$ & $0.181$ & $0.215$ & $1.19\times$ \\
O-LoRA  & $0.840$ & $0.990$ & $1.18\times$ & $0.213$ & $0.337$ & $1.58\times$ \\
ELLA    & $0.462$ & $0.954$ & $2.06\times$ & $0.341$ & $0.893$ & $2.62\times$ \\
\bottomrule
\end{tabular}
\end{table}

\begin{table}[t]
\begin{minipage}[t]{0.405\textwidth}
\centering
\footnotesize
\caption{\footnotesize Update concentration $\sigma_1/\lVert \Delta W \rVert_F$
 and effective directions $1/\text{ratio}^2$,
median over tracked runs and logged steps. Standard CL, $15$ runs per
cell. Flat-spectrum reference $0.354$ ($8$ directions).}
\label{tab:concentration}
\setlength{\tabcolsep}{3pt}
\begin{tabular}{lcccc}
\toprule
& \multicolumn{2}{c}{AdamW}& \multicolumn{2}{c}{Muon}\\
\cmidrule(lr){2-3} \cmidrule(lr){4-5}
& ratio & dir. & ratio & dir. \\
\midrule
IncLoRA & $0.854$ & $1.37$ & $0.378$ & $6.99$ \\
O-LoRA  & $0.838$ & $1.42$ & $0.378$ & $6.99$ \\
ELLA    & $0.737$ & $1.84$ & $0.379$ & $6.95$ \\
\bottomrule
\end{tabular}
\end{minipage}
\hfill
\begin{minipage}[t]{0.575\textwidth}
\centering
\footnotesize
\caption{\footnotesize Mean Muon $-$ AdamW difference within each
$(\text{seed}, \text{order})$ cell, with the number of cells sharing its sign
($15$ on Standard CL, $3$ on TRACE). On TRACE the $\Delta$FWT count is of
\emph{negative} cells.}
\label{tab:paired}
\setlength{\tabcolsep}{3pt}
\begin{tabular}{lcccc}
\toprule
& \multicolumn{2}{c}{Standard CL} & \multicolumn{2}{c}{TRACE} \\
\cmidrule(lr){2-3} \cmidrule(lr){4-5}
& $\Delta$OA & $\Delta$FWT & $\Delta$OA & $\Delta$FWT \\
\midrule
IncLoRA & $+5.41$ \tiny$12/15$ & $+18.19$ \tiny$15/15$ & $+5.44$ \tiny$3/3$ & $-2.93$ \tiny$3/3$ \\
O-LoRA  & $-0.72$ \tiny$5/15$  & $+18.30$ \tiny$15/15$ & $+7.62$ \tiny$3/3$ & $-3.46$ \tiny$3/3$ \\
ELLA    & $-8.40$ \tiny$0/15$  & $+9.94$ \tiny$10/15$  & $+3.42$ \tiny$3/3$ & $-1.87$ \tiny$3/3$ \\
\bottomrule
\end{tabular}
\end{minipage}
\end{table}

\section{Discussion}
\label{sec:discussion}

Our two mechanistic measurements suggest an account of why the optimizer and an
explicit penalty do not combine. The penalties act on the magnitude of the
update (Table~\ref{tab:magnitude}), while Muon fixes that magnitude by
construction and instead distributes the update across singular directions
(Table~\ref{tab:concentration}). A penalty therefore has less to act on once
Muon is in place, and what remains of it changes the direction of the update
without the accompanying reduction in scale. On Standard CL the cost of stacking
follows the ordering of the magnitudes the penalties produce under AdamW, and
the fit decomposition makes that cost concrete: ELLA${+}$Muon retains as well as
any other configuration while fitting each task least well of the six, despite
taking the largest updates of any ELLA configuration.%

Why distributing the update helps at all, we do not establish. One reading is
that a network whose updates occupy many directions adapts its features rather
than learning with a fixed kernel \citep{chizat2019lazy}, which would be
consistent with transfer moving away from zero in both directions rather than
being suppressed toward it. Testing that requires measuring how far the features
move during adaptation, which we have not done.%

\paragraph{What we do not claim.}
(i) The ranking among the best configurations is not resolved by our data: on
Standard CL the unconstrained baseline with Muon leads the best AdamW
configuration by $0.24$ points, in $7$ of $15$ paired cells, and on TRACE the
comparison reverses.
(ii) The concentration ratio measures the distribution of a \emph{single} update
across singular directions; it does not show that \emph{successive} task updates
use different directions, which is what O-LoRA constrains. Principal angles
between consecutive adapters would settle that and we have not measured them.
(iii) The step-size matching is applied on TRACE only, and with the
$\mathcal{O}(1)$ constant $c$ of Appendix~\ref{app:scaling-equivalence} set to
one. We do not measure $c$: our Standard CL magnitudes are norms of the merged
product rather than of the factors the optimizers update, so they do not bear on
it. The Standard CL comparison, in turn, is not matched by construction but
measured after the fact (Table~\ref{tab:magnitude}).
(iv) The two benchmarks are unequally sampled ($15$ paired cells per
configuration on Standard CL against $3$ seeds on one task order on TRACE) and
they do not agree: the grouping by number of mechanisms orders the six
configurations on Standard CL but not on TRACE, where the optimizer alone
separates them. We cannot say whether that reflects the benchmark or the
protocol, which differ in both respects.
(v) We use O-LoRA and ELLA at the hyperparameters their authors report, tuned
under AdamW, and did not re-tune their penalty strengths under Muon; the
ELLA${+}$Muon result is a statement about the combination as it comes, not about
every setting of the penalty.
(vi) The alignment of each task sequence is taken from the benchmark
descriptions rather than measured.
(vii) Everything here is a text encoder-decoder at the Large scale; whether the
effect survives at generative LLM scale, or on vision and multimodal backbones,
is open.

\section{Future Work}
\label{sec:futurework}

First, we will measure the principal angles between the adapters of consecutive
tasks, comparing IncLoRA${+}$Muon against O-LoRA${+}$AdamW. This is the missing
link between our concentration measurement and the property constraint-based
methods impose: if updates trained with Muon turn out to be as mutually
orthogonal as those O-LoRA constrains to be, the substitution argument closes.
Second, we will measure the constant $c$ of
Appendix~\ref{app:scaling-equivalence} per adapter matrix and re-run TRACE at
the measured value, since the rule as applied assumes $c=1$ without
verification.
Third, we will run more task orders on TRACE, where the grouping by number of
mechanisms does not reproduce what we see on Standard CL, and extend the
evaluation to generative language models to test whether the effect of Muon's
update geometry holds at scale.
Finally, the matching rule equalizes the \emph{nominal} step, not the realized
one. Having separated the scale of the update from its distribution across
singular directions (Table~\ref{tab:magnitude} from
Table~\ref{tab:concentration}), we want to test whether the distribution alone
is sufficient, by constructing an optimizer that flattens the spectrum while
leaving the realized step norm free, and one that does the reverse.%

\bibliography{collas2026_conference}
\bibliographystyle{collas2026_conference}

\newpage
\appendix

\section{Protocol}
\label{app:protocol}
Table~\ref{tab:protocol} lists the two settings. We evaluate the concentration
ratio every tenth optimizer step and the relative Frobenius norm of
Appendix~\ref{formula:frobenius-norm} every fifth, per module.%

\begin{table}[ht]
\centering
\footnotesize
\caption{Protocol. The two settings differ in backbone, budget and whether
the nominal step sizes are matched; everything else is held fixed within a
setting.}
\label{tab:protocol}
\begin{tabular}{lll}
\toprule
& Standard CL & TRACE \\
\midrule
Backbone                  & T5-Large & Flan-T5-Large \\
Tasks $K$                 & $4$ & $8$ \\
Task orders               & Orders 1, 2, 3 & Order 7 \\
Seeds                     & $42$, $121$, $1234$, $1337$, $3407$ & $42$, $121$, $3407$ \\
Runs per configuration    & $15$ ($90$ in total) & $3$ ($18$ in total) \\
LoRA rank / $\alpha$ / targets & $8$ / $16$ / query, value & $8$ / $16$ / query, value \\
Optimizer steps per run   & $821$ & $\sim3700$ (early stopping) \\
AdamW learning rate      & $1\mathrm{e}{-3}$ (ELLA protocol) & $1\mathrm{e}{-5}$ \\
Muon learning rate        & $1\mathrm{e}{-3}$ (\emph{not} matched) & $3.2\mathrm{e}{-4}$ \\
Training budget           & fixed epochs (ELLA protocol) & early stopping on validation loss (patience $3$) \\
\bottomrule
\end{tabular}
\end{table}

\section{Relative Frobenius norm}
\label{formula:frobenius-norm}
The relative Frobenius norm quantifies the aggregated magnitude of the
adapter update across all $L$ adapter modules, at training step $t$:

\[
\text{Global Rel-Fro}_t \;=\; \sqrt{ \frac{\sum_{l=1}^{L} s^2 \big\| B_t^{(l)}A_t^{(l)} - B_{t-1}^{(l)}A_{t-1}^{(l)} \big\|_F^2}{\sum_{l=1}^{L} \big\| W_0^{(l)} \big\|_F^2} }
\]

where $\|\cdot\|_F$ denotes the Frobenius norm, $s = \alpha/r$ is the LoRA
scaling, $B_t^{(l)}A_t^{(l)}$ is the merged adapter product of the $l$-th module
at step $t$, and $W_0^{(l)}$ is the frozen backbone weight of that module. The
denominator is therefore constant in $t$, so the quantity is the magnitude of a
single optimizer step normalized by a fixed reference rather than by the current
adapter state. Note also that the numerator is not the matrix either optimizer
updates, which is each LoRA factor separately, so these ratios do not convert
into the constant $c$ of Appendix~\ref{app:scaling-equivalence}.

\section{Update concentration}
\label{app:concentration}

The concentration ratio is computed per adapter module, on the $144$ query and
value projections of T5-Large. For module $l$ at step $t$ we form the merged
update
\[
    \Delta W_t^{(l)} \;=\; B_t^{(l)} A_t^{(l)} \;-\; B_{t-1}^{(l)} A_{t-1}^{(l)},
\]
take its singular values, and evaluate $\sigma_1 / \lVert \Delta W_t^{(l)}
\rVert_F$. The two states are one optimizer step apart, not one logging interval.
We aggregate by median across modules rather than pooling them into one global
ratio, because a global $\sqrt{\sum_l \lVert \cdot \rVert_F^2}$ over a single
$\max_l \sigma_1$ mixes the spectra of matrices of different sizes. Steps are
pooled within a run before aggregating across runs, so that runs contribute
equally regardless of length.

Two properties make this quantity convenient. It is dimensionless, so it is
unaffected by the LoRA scaling $\alpha/r$ and by any learning-rate choice; and
its range is bounded, so a measured value can be compared against a
flat-spectrum value rather than only against another method. Because
$\Delta W_t^{(l)}$ is a difference of two rank-$r$ products its rank is at most
$2r$, so the ratio cannot fall below $1/\sqrt{2r} = 0.25$ for $r = 8$.
Empirically the spectrum is effectively of rank $r$: fewer than $1\%$ of the
update's energy lies beyond the first $r$ singular values under either optimizer
(median $0.80\%$ under AdamW, $0.79\%$ under Muon), and under Muon there is a
sharp drop at rank $r$ (median $\sigma_{r+1}/\sigma_r \approx 0.13$, against
$\approx 0.90$ under AdamW). The corresponding flat-spectrum value,
$1/\sqrt{r} = 0.354$, is also what Muon's construction imposes on the update to
each factor, which is what makes it the right value to compare the product
against.

These two figures are reconstructed from $64$-bin histograms of the singular
values, logged every twentieth step for the $64$ of $90$ runs whose spectra are
available locally; the concentration ratio itself is logged directly as a scalar
for all $90$.

Within a task the median ratio is $0.847$ (AdamW) and $0.378$ (Muon); in a
$\pm 10$-step window around the three task boundaries of Order~1 it is $0.813$
and $0.379$, so the separation between the optimizers is not an artefact of task
transitions. The tracker re-snapshots the previous state when a new adapter
is initialized, so the boundary spikes of Figure~\ref{fig:update_magnitude} are
genuine first steps on a fresh adapter rather than an artefact of the
measurement.

\section{Equivalence of the Matched Learning Rate}
\label{app:scaling-equivalence}

We justify the scaling rule $\eta_{\text{Muon}} = \eta_{\text{AdamW}}\,\sqrt{\max(m,n)}$
used in Section~\ref{sec:methods}. The goal is to pick $\eta_{\text{Muon}}$ so that
a single Muon step perturbs the entries of an adapter matrix by the same
root-mean-square (RMS) amount as an AdamW step taken at learning rate
$\eta_{\text{AdamW}}$. Matching the per-entry update magnitude isolates the
\emph{direction} of the update from its scale.

For a matrix $A \in \mathbb{R}^{m \times n}$ we use the entrywise RMS
\begin{equation}
    \mathrm{RMS}(A) \;=\; \sqrt{\frac{1}{mn}\sum_{i,j} A_{ij}^2}
    \;=\; \frac{\lVert A \rVert_F}{\sqrt{mn}}.
\end{equation}

\paragraph{AdamW update.} The AdamW update of a weight matrix $W \in \mathbb{R}^{m\times n}$ is
$\Delta W_{\text{AdamW}} = \eta_{\text{AdamW}}\, \big(\hat{m} \oslash (\sqrt{\hat{v}} + \epsilon)\big)$,
where $\oslash$ denotes entrywise division and $\hat{m}, \hat{v}$ are the bias-corrected
first and second moment estimates. Under the standard idealization that each
coordinate is normalized by its own running scale, i.e.\ $\hat{m}_{ij}/\sqrt{\hat{v}_{ij}} \approx \pm 1$,
every entry of the update direction has unit magnitude, so
\begin{equation}
    \mathrm{RMS}(\Delta W_{\text{AdamW}}) \;\approx\; \eta_{\text{AdamW}}.
    \label{eq:adamw-rms}
\end{equation}
More generally $\mathrm{RMS}(\Delta W_{\text{AdamW}}) = c\,\eta_{\text{AdamW}}$ for some
$\mathcal{O}(1)$ constant $c$; we take $c = 1$ and absorb it into the base learning rate.

\paragraph{Muon update.} Muon replaces the momentum $G = U\Sigma V^\top$ by its
orthogonalization $O = \mathrm{NewtonSchulz}(G) \approx U V^\top$ and updates
$\Delta W_{\text{Muon}} = \eta_{\text{Muon}}\, O$. Since $U$ and $V$ have orthonormal
columns, $O$ has $r = \min(m,n)$ singular values all equal to one, so
\begin{equation}
    \lVert O \rVert_F^2 \;=\; \mathrm{tr}(O^\top O)
    \;=\; \mathrm{tr}\!\big(V U^\top U V^\top\big)
    \;=\; \mathrm{tr}\!\big(V V^\top\big)
    \;=\; r \;=\; \min(m,n).
\end{equation}
Using $\min(m,n)\cdot \max(m,n) = mn$, the update magnitude is therefore
\begin{equation}
    \mathrm{RMS}(\Delta W_{\text{Muon}})
    \;=\; \eta_{\text{Muon}}\,\frac{\lVert O \rVert_F}{\sqrt{mn}}
    \;=\; \eta_{\text{Muon}}\,\frac{\sqrt{\min(m,n)}}{\sqrt{mn}}
    \;=\; \frac{\eta_{\text{Muon}}}{\sqrt{\max(m,n)}}.
    \label{eq:muon-rms}
\end{equation}

\paragraph{Matching the two.} Equating the update magnitudes
\eqref{eq:adamw-rms} and \eqref{eq:muon-rms},
\begin{equation}
    \frac{\eta_{\text{Muon}}}{\sqrt{\max(m,n)}} \;=\; \eta_{\text{AdamW}}
    \quad\Longleftrightarrow\quad
    \eta_{\text{Muon}} \;=\; \eta_{\text{AdamW}}\,\sqrt{\max(m,n)},
\end{equation}
which is the rule used in the main text. \hfill$\square$

\paragraph{Remarks.} This recovers the per-parameter update-scale rule of
\citet{liu2025muonscalable}, derived for large-scale pretraining and applied here
per adapter matrix, so $(m,n)$ are the dimensions of the matrix being updated. The
derivation assumes zero weight decay ($\lambda = 0$) for both optimizers, so no
decoupled $-\eta\lambda W$ term enters the RMS balance.
This holds on Standard CL, where both optimizers run at $\lambda = 0$; on
TRACE AdamW uses $\lambda = 0.01$, which the balance above neglects.
We take $c = 1$ without measuring it. Our Standard CL runs do not settle the
matter: there both optimizers use the same learning rate, and for the
$1024$-dimensional query and value projections $c = 1$ would predict a factor of
$\sqrt{\max(m,n)} = 32$ between the two per-entry update magnitudes, whereas the
realized magnitudes we measure differ by at most a factor of about two
(Table~\ref{tab:magnitude}). Those magnitudes are computed on the merged product
$BA$ rather than on the factors the optimizers update
(Appendix~\ref{formula:frobenius-norm}), and the update to the product carries
the factor norms as well as the step size, so the discrepancy does not convert
into a value of $c$. We list the direct measurement as future work
(Section~\ref{sec:futurework}). Because the concentration ratio is
dimensionless, none of the results in Table~\ref{tab:concentration} depend on
the choice of $c$.

\end{document}

%% file: math_commands.tex
\usepackage{amsmath,amsfonts,bm}

\def\eqref#1{equation~\ref{#1}}

\def\1{\bm{1}}

\DeclareMathAlphabet{\mathsfit}{\encodingdefault}{\sfdefault}{m}{sl}
\SetMathAlphabet{\mathsfit}{bold}{\encodingdefault}{\sfdefault}{bx}{n}

